\documentclass[journal]{IEEEtran}
\usepackage{amsmath}
\usepackage{amsfonts}
\usepackage{multirow}
\usepackage{array}
 \usepackage{graphicx}
\begin{document}

\title{Evaluating Word Embedding Models: Methods and Experimental Results}

\author{Bin~Wang$^{*}$, \IEEEmembership{Student Member, IEEE},~Angela~Wang$^{*}$,~Fenxiao~Chen, \IEEEmembership{Student Member, IEEE},\\~Yuncheng~Wang~
and~C.-C.~Jay~Kuo,~\IEEEmembership{Fellow, IEEE}
\thanks{
Bin Wang, Fenxiao Chen, Yunchen Wang and C.-C. Jay Kuo are with the Ming-Hsieh Department of 
Electrical and Computer Engineering, University of Southern California, Los Angeles, CA 90089
-2564, USA. Email: \{wang699, fenxiaoc\}@usc.edu, cckuo@sipi.usc.edu}
\thanks{
Angela Wang is with the Department of Electrical Engineering and Computer Science, 
University of California, Berkeley, Berkeley, CA 94720-2278, USA. Email: awangs@berkeley.edu}
\thanks{
$^*$ These authors contributes equally to this work.}
}

%\markboth{Journal of \LaTeX\ Class Files,~Vol.~14, No.~8, August~2018}%

\maketitle

\begin{abstract}

Extensive evaluation on a large number of word embedding models for
language processing applications is conducted in this work.  First, we
introduce popular word embedding models and discuss desired properties
of word models and evaluation methods (or evaluators).  Then, we
categorize evaluators into intrinsic and extrinsic two types.  Intrinsic
evaluators test the quality of a representation independent of specific
natural language processing tasks while extrinsic evaluators use word
embeddings as input features to a downstream task and measure changes in
performance metrics specific to that task. We report experimental
results of intrinsic and extrinsic evaluators on six word embedding
models.  It is shown that different evaluators focus on different
aspects of word models, and some are more correlated with natural
language processing tasks.  Finally, we adopt correlation analysis to
study performance consistency of extrinsic and intrinsic evalutors. 

\end{abstract}

\begin{IEEEkeywords}
Word embedding, Word embedding evaluation, Natural language processing.
\end{IEEEkeywords}

\IEEEpeerreviewmaketitle

\section{Introduction}\label{sec:introduction}

\IEEEPARstart{W}{ord} embedding is a real-valued vector representation
of words by embedding both semantic and syntactic meanings obtained from
unlabeled large corpus. It is a powerful tool widely used in modern
natural language processing (NLP) tasks, including semantic analysis
\cite{yu2018refining}, information retrieval
\cite{schutze2008introduction}, dependency parsing
\cite{chen2015distributed, ouchi2016transition, shen2014dependency},
question answering \cite{zhou2016learning, hao2017end} and machine
translation \cite{zhou2016learning, zhang2017context, chen2018neural}.
Learning a high quality representation is extremely important for these
tasks, yet the question ``what is a good word embedding model'' remains
an open problem. 

Various evaluation methods (or evaluators) have been proposed to test
qualities of word embedding models.  As introduced in \cite{survey}, there are
two main categories for evaluation methods -- intrinsic and extrinsic
evaluators. Extrinsic evaluators use word embeddings as input features
to a downstream task and measure changes in performance metrics specific
to that task. Examples include part-of-speech tagging
\cite{li2014joint}, named-entity recognition \cite{xu2018cross},
sentiment analysis \cite{ravi2015survey} and machine translation
\cite{bahdanau2014neural}. Extrinsic evaluators are more computationally
expensive, and they may not be directly applicable.  Intrinsic
evaluators test the quality of a representation independent of specific
natural language processing tasks. They measure syntactic or semantic
relationships among words directly. Aggregate scores are given from
testing the vectors in selected sets of query terms and semantically
related target words. One can further classify intrinsic evaluators into
two types: 1) absolute evaluation, where embeddings are evaluated
individually and only their final scores are compared, and 2)
comparative evaluation, where people are asked about their preferences
among different word embeddings \cite{eval_methods}. Since comparative
intrinsic evaluators demand additional resources for subjective tests,
they are not as popular as the absolute ones. 

A good word representation should have certain good properties.  An
ideal word evaluator should be able to analyze word embedding models
from different perspectives. Yet, existing evaluators put emphasis on a
certain aspect with or without consciousness.  There is no unified
evaluator that analyzes word embedding models comprehensively.
Researchers have a hard time in selecting among word embedding models
because models do not always perform at the same level on different
intrinsic evaluators. As a result, the gold standard for a good word
embedding model differs for different language tasks. In this work, we
will conduct correlation study between intrinsic evaluators and language
tasks so as to provide insights into various evaluators and help
people select word embedding models for specific language tasks. 

Although correlation between intrinsic and extrinsic evaluators was
studied before \cite{chiu2016intrinsic, qiu2018revisiting}, this topic
is never thoroughly and seriously treated. For example, producing models
by changing the window size only does not happen often in real world
applications, and the conclusion drawn in \cite{chiu2016intrinsic} might
be biased.  The work in \cite{qiu2018revisiting} only focused on Chinese
characters with limited experiments. We provide the most comprehensive
study and try to avoid the bias as much as possible in this work.

The rest of the paper is organized as follows.  Popular word embedding
models are reviewed in Sec. \ref{sec:review}.  Properties of good
embedding models and intrinsic evaluators are discussed in Sec.
\ref{sec:properties}. Representative performance metrics of intrinsic
evaluation are presented in Sec. \ref{sec:IE} and the corresponding
experimental results are offered in Sec. \ref{sec:exp_IE}.
Representative performance metrics of extrinsic evaluation are
introduced in Sec. \ref{sec:EE} and the corresponding experimental
results are provided in Sec.  \ref{sec:exp_EE}. We conduct consistency
study on intrinsic and extrinsic evaluators using correlation analysis
in Sec.  \ref{sec:correlation}.  Finally, concluding remarks and future
research directions are discussed in Sec.  \ref{sec:future}. 

\section{Word Embedding Models}\label{sec:review}

As extensive NLP downstream tasks emerge, the demand for word embedding is growing significantly. As a result, lots of word embedding methods are proposed while some of them share the same concept. We categorize the existing word embedding methods based on their techniques. % A number of word embedding models are summarized below. 

\subsection{Neural Network Language Model (NNLM)}

The Neural Network Language Model (NNLM) \cite{nnlm} jointly learns a
word vector representation and a statistical language model with a
feedforward neural network that contains a linear projection layer and a
non-linear hidden layer. An $N$-dimensional one-hot vector that
represents the word is used as the input, where $N$ is the size of the
vocabulary.  The input is first projected onto the projection layer.
Afterwards, a softmax operation is used to compute the probability
distribution over all words in the vocabulary.  As a result of its
non-linear hidden layers, the NNLM model is very computationally
complex. To lower the complexity, an NNLM is first trained using
continuous word vectors learned from simple models. Then, another N-gram
NNLM is trained from the word vectors. 

\subsection{Continuous-Bag-of-Words (CBOW) and Skip-Gram}
	
Two iteration-based methods were proposed in the word2vec paper
\cite{word2vec}. The first one is the Continuous-Bag-of-Words (CBOW)
model, which predicts the center word from its surrounding context.
This model maximizes the probability of a word being in a specific
context in form of
\begin{equation}
P(w_i|w_{i-c},w_{i-c+1},...,w_{i-1},w_{i+1},..,w_{i+c-1},w_{i+c}),
\end{equation}
where $w_i$ is a word at position $i$ and $c$ is the window size. Thus,
it yields a model that is contingent on the distributional similarity of
words. 

We focus on the first iteration in the discussion below.  Let $W$ be the
vocabulary set containing all words.  The CBOW model trains two
matrices: 1) an input word matrix denoted by $V \in \mathbb{R}^{N \times
|W|}$, where the $i^{th}$ column of $V$ is the $N$-dimensional embedded
vector for input word $v_i$, and 2) an output word matrix denoted by $U
\in \mathbb{R}^{|W|\times N}$, where the $j^{th}$ row of $U$ is the
$N$-dimensional embedded vector for output word $u_j$. To embed input
context words, we use the one-hot representation for each word
initially, and apply $V^{T}$ to get the corresponding word vector
embeddings of dimension $N$. We apply $U^{T}$ to an input word vector to
generate a score vector and use the softmax operation to convert a score
vector into a probability vector of size $W$. This process is to yield a
probability vector that matches the vector representation of the output
word.  The CBOW model is obtained by minimizing the cross-entropy loss
between the probability vector and the embedded vector of the output
word.  This is achieved by minimizing the following objective function:
\begin{equation}
J (u_i) = -u^T_i \hat{v} + \log \sum_{j=1}^{|W|} \exp(u^T_j \hat{v}), 
\end{equation} 
where $u_i$ is the $i^{th}$ row of matrix $U$ and $\hat{v}$ is the
average of embedded input words. 

Initial values for matrices $V$ and $U$ are randomly assigned. The
dimension $N$ of word embedding can vary based on different application
scenarios. Usually, it ranges from $50$ to $300$ dimensions. After
obtaining both matrices $V$ or $U$, they can either be used solely or
averaged to obtained the final word embedding matrix. 
     
The skip-gram model \cite{word2vec} predicts the surrounding context
words given a center word. It focuses on maximizing probabilities of
context words given a specific center word, which can be written as
\begin{equation}
P(w_{i-c},w_{i-c+1},...,w_{i-1},w_{i+1},..,w_{i+c-1},w_{i+c}|w_i).
\end{equation}
The optimization procedure is similar to that for the CBOW model but
with a reversed order for context and center words. 
  
The softmax function mentioned above is a method to generate probability
distributions from word vectors. It can be written as
\begin{equation}
P(w_c|w_i) = \frac{\exp(v_{w_c}^T v_{w_i})}{\sum_{w=1}^{|W|} 
\exp(v_{w}^T v_{w_i})}.
\end{equation}
This softmax function is not the most efficient one since we must take a
sum over all $W$ words to normalize this function. Other functions that
are more efficient include negative sampling and hierarchical softmax
\cite{softmax}. Negative sampling is a method that maximizes the log
probability of the softmax model by only summing over a smaller subset
of $W$ words. Hierarchical softmax also approximates the full softmax
function by evaluating only $\log_2 W$ words. Hierarchical softmax uses a
binary tree representation of the output layer where the words are
leaves and every node represents the relative probabilities of its child
nodes. These two approaches do well in making predictions for local
context windows and capturing complex linguistic patterns. Yet, it could
be further improved if global co-occurrence statistics is leveraged. 

\subsection{Co-occurrence Matrix}

In our current context, the co-occurrence matrix is a word-document
matrix. The $(i,j)$ entry, $X_{ij}$, of co-occurrence matrix ${\bf X}$
is the number of times for word $i$ in document $j$. This definition can
be generalized to a window-based co-occurence matrix where the number of
times of a certain word appearing in a specific sized window around a
center word is recorded.  In contrast with the window-based log-linear
model representations (e.g. CBOW or Skip-gram) that use local
information only, the global statistical information is exploited by
this approach. 
        
One method to process co-occurrence matrices is the singular value
decomposition (SVD). The co-occurrence matrix is expressed in form of
$USV^T$ matrices product, where the first $k$ columns of both $U$ and
$V$ are word embedding matrices that transform vectors into a
$k$-dimensional space with an objective that it is sufficient to capture
semantics of words. Although embedded vectors derived by this procedure
are good at capturing semantic and syntactic information, they still
face problems such as imbalance in word frequency, sparsity and high
dimensionality of embedded vectors, and computational complexity. 
        
To combine benefits from the SVD-based model and the log-linear models,
the Global Vectors (GloVe) method \cite{glove} adopts a weighted
least-squared model. It has a framework similar to that of the skip-gram
model, yet it has a different objective function that contains
co-occurence counts. We first define a word-word co-occurence matrix
that records the number of times word $j$ occurs in the context of word
$i$.  By modifying the objective function adopted by the skip-gram
model, we derive a new objective function in form of
\begin{equation}
\hat{J} = \sum_{i=1}^{W} \sum_{j=1}^{W} f(X_{ij})(u_j^T v_i - \log X_{ij})^2,
\end{equation}
where $f(X_{ij})$ is the number of times word $j$ occurs in the context of word $i$.

The GloVe model is more efficient as its objective function contains
nonzero elements of the word-word co-occurrence matrix only. Besides,
it produces a more accurate result as it takes co-occurrence counts
into account. 

\subsection{FastText}
	
Embedding of rarely used words can sometimes be poorly estimated. Therefore
several methods have been proposed to remedy this issue, including the FastText
method. FastText uses the subword information explicitly so embedding for rare words
can still be represented well. It is still based on the skip-gram model,
where each word is represented as a bag of character $n$-grams or
subword units \cite{fasttext}. A vector representation is associated
with each of character $n$-grams, and the average of these vectors gives the
final representation of the word.  This model improves the performance
on syntactic tasks significantly but not much in semantic questions. 
	
\subsection{N-gram Model}

The N-gram model is an important concept in language models. It has been
used in many NLP tasks. The ngram2vec method \cite{zhao2017ngram2vec}
incorporates the n-gram model in various baseline embedding models such
as word2vec, GloVe, PPMI and SVD.  Furthermore, instead of using
traditional training sample pairs or the sub-word level information such
as FastText, the ngram2vec method considers word-word level
co-occurrence and enlarges the reception window by adding the word-ngram
and the ngram-ngram co-occurrence information. Its performance on word
analogy and word similarity tasks has significantly improved. It is also
be able to learn negation word pairs/phrases like 'not interesting',
which is a difficult case for other models. 

\subsection{Dictionary Model}

Even with larger text data available, extracting and embedding all
linguistic properties into a word representation directly is a
challenging task. Lexical databases such as the WordNet are helpful to
the process of learning word embeddings, yet labeling large lexical
databases is a time-consuming and error-prone task. In contrast, a
dictionary is a large and refined data source for describing words. The
dict2vec method learns word representation from dictionary entries as
well as large unlabeled corpus \cite{tissier2017dict2vec}.  Using the
semantic information from a dictionary, semantically-related words tend
to be closer in high-dimensional vector space. Also, negative sampling
is used to filter out pairs which are not correlated in a dictionary. 
	
\subsection{Deep Contextualized Model}

To represent complex characteristics of words and word usage across
different linguistic contexts effectively, a new model for deep
contextualized word representation was introduced in \cite{deep}. First,
an Embeddings from Language Models (ELMo) representation is generated
with a function that takes an entire sentence as the input. The function
is generated by a bidirectional LSTM network that is trained with a
coupled language model. Existing embedding models can be improved by
incorporating the ELMo representation as it is effective in
incorporating the sentence information.  By following ELMo, a series of
pre-trained neural network models for language tasks are proposed such as BERT
\cite{devlin2018bert} and OpenAI GPT \cite{radford2018improving}. Their effectiveness 
is proved in lots of language tasks.

\section{Desired Properties of Embedding Models and Evaluators}\label{sec:properties}

\subsection{Embedding Models}\label{WE_Property}
 
Different word embedding models yield different vector representations.
There are a few properties that all good representations should aim for. 

\begin{itemize}
\item \textit{Non-conflation} \cite{props}

Different local contexts around a word should give rise to specific
properties of the word, e.g., the plural or singular form, the tenses,
etc. Embedding models should be able to discern differences in the
contexts and encode these details into a meaningful representation in
the word subspace. 
        
\item \textit{Robustness Against Lexical Ambiguity} \cite{props}

All senses (or meanings) of a word should be represented. Models should
be able to discern the sense of a word from its context and find the
appropriate embedding. This is needed to avoid meaningless
representations from conflicting properties that may arise from the
polysemy of words. For example, word models should be able to represent
the difference between the following: ``the \textbf{bow} of a ship" and
``\textbf{bow} and arrows". 
        
\item \textit{Demonstration of Multifacetedness} \cite{props}

The facet, phonetic, morphological, syntactic, and other properties, of
a word should contribute to its final representation.  This is important
as word models should yield meaningful word representations and perhaps
find relationships between different words.  For example, the
representation of a word should change when the tense is changed or a
prefix is added. 
        
\item \textit{Reliability} \cite{reliability}

Results of a word embedding model should be reliable. This is important
as word vectors are randomly initialized when being trained.  Even if a
model creates different representations from the same dataset because of
random initialization, the performance of various representations should
score consistently. 
        
\item \textit{Good Geometry} \cite{goodgeo}

The geometry of an embedding space should have a good spread. Generally
speaking, a smaller set of more frequent, unrelated words should be
evenly distributed throughout the space while a larger set of rare words
should cluster around frequent words. Word models should overcome the
difficulty arising from inconsistent frequency of word usage and derive
some meaning from word frequency. 

\end{itemize}
    
\subsection{Evaluators}\label{IE_Property}

The goal of an evaluator is to compare characteristics of different word
embedding models with a quantitative and representative metric.
However, it is not easy to find a concrete and uniform way in evaluating
these abstract characteristics. Generally, a good word embedding
evaluator should aim for following properties. 
    
\begin{itemize}

\item \textit{Good Testing Data}

To ensure a reliable representative score, testing data should be varied
with a good spread in the span of a word space. Frequently and rarely
occurring words should be included in the evaluation. Furthermore, data
should be reliable in the sense that they are correct and objective. 

\item \textit{Comprehensiveness} 

Ideally, an evaluator should test for many properties of a word
embedding model.  This is not only an important property for giving a
representative score but also for determining the effectiveness of an
evaluator. 
            
\item \textit{High correlation} 

The score of a word model in an intrinsic evaluation task should
correlate well with the performance of the model in downstream natural
language processing tasks. This is important for determining the
effectiveness of an evaluator. 
            
\item \textit{Efficiency} 

Evaluators should be computationally efficient.  Most models are created
to solve computationally expensive downstream tasks. Model evaluators
should be simple yet able to predict the downstream performance of a
model. 
        
\item \textit{Statistical Significance} 

The performance of different word embedding models with respect to an
evaluator should have enough statistical significance, or enough
variance between score distributions, to be differentiated
\cite{statsig}.  This is needed in judging whether a model is better
than another and helpful in determining performance rankings between
models. 

\end{itemize}

\section{Intrinsic Evaluators}\label{sec:IE}

Intrinsic evaluators test the quality of a representation independent of specific natural language processing tasks. They measure syntactic or semantic relationships between word directly. In this section, a number of absolute intrinsic evaluators will be discussed. 
    
\subsection{Word Similarity} 

The word similarity evaluator correlates the distance between word
vectors and human perceived semantic similarity. The goal is to measure
how well the notion of human perceived similarity is captured by the
word vector representations, and validate the distributional hypothesis
where the meaning of words is related to the context they occur in. For
the latter, the way distributional semantic models simulate similarity
is still ambiguous \cite{ws}. 
        
One commonly used evaluator is the cosine similarity defined by
\begin{equation}
\cos(w_x,w_y) = \frac{w_x \cdot w_y}{||w_x|| \ ||w_y||},
\end{equation}
where $w_x$ and $w_y$ are two word vectors and $||w_x||$ and $||w_y||$
are the $\ell_2$ norm.  This test computes the correlation between all
vector dimensions, independent of their relevance for a given word pair
or for a semantic cluster. 

Because its scores are normalized by the vector length, it is robust to
scaling. It is computationally inexpensive. Thus, it is easy to compare
multiple scores from a model and can be used in word model's prototyping
and development.  Furthermore, word similarity can be used to test
model's robustness against lexical ambiguity, as a dataset aimed at
testing multiple senses of a word can be created. 
        
On the other hand, it has several problems as discussed in \cite{ws}.
This test is aimed at finding the distributional similarity among pairs
of words, but this is often conflated with morphological relations and
simple collocations. Similarity may be confused with relatedness. For
example, $car$ and $train$ are two similar words while $car$ and $road$ are two
related words. The correlation between the score from the intrinsic test
and other extrinsic downstream tasks could be low in some cases.  There
is doubt about the effectiveness of this evaluator because it might not
be comprehensive. 
    
\subsection{Word Analogy}

When given a pair of words $a$ and $a^*$ and a third word $b$, the
analogy relationship between $a$ and $a^*$ can be used to find the
corresponding word $b^*$ to $b$. Mathematically, it is expressed as
\begin{equation}
a:a^* :: b: \_\_,
\end{equation}
where the blank is $b^*$. One example could be 
\begin{equation}
\mbox{write}:\mbox{writing} :: \mbox{read}: \mbox{\underline{reading}}. 
\end{equation}
The 3CosAdd method \cite{3cosadd} solves for $b^*$ using the following equation:
\begin{equation}
b^* = \underset{b'}{\mathrm{argmax}} (\cos(b', a^*-a+b)),
\end{equation}
Thus, high cosine similarity means that vectors share a similar
direction. However, it is important to note that the 3CosAdd
method normalizes vector lengths using the cosine similarity
\cite{3cosadd}. Alternatively, there is the 3CosMul
\cite{pairdir} method, which is defined as
\begin{equation}
b^* = \underset{b'}{\mathrm{argmax}} \dfrac{\cos(b',b) \cos(b',a^*)}{\cos(b',a)+\varepsilon}
\end{equation}
where $\varepsilon=0.001$ is used to prevent division by zero.  The
3CosMul method has the same effect with taking the logarithm of each
term before summation.  That is, small differences are enlarged while
large ones are suppressed.  Therefore, it is observed that the 3CosMul
method offers better balance in different aspects. 
      
It was stated in \cite{analogy_issues} that many models score under 30\%
on analogy tests, suggesting that not all relations can be identified in
this way. In particular, lexical semantic relations like synonymy and
antonym are the most difficult. They also concluded that the analogy test is
the most successful when all three source vectors are relatively close to
the target vector. Accuracy of this test decreases as their distance
increases. Another seemingly counter-intuitive finding is that words
with denser neighborhoods yield higher accuracy. This is perhaps because
of its correlation with distance. Another problem with this test is
subjectivity. Analogies are fundamental to human reasoning and logic.
The dataset on which current word models are trained does not encode our
sense of reasoning. It is rather different from the way how humans learn
natural languages. Thus, given a word pair, the vector space model may
find a different relationship from what humans may find. 
        
Generally speaking, this evaluator serves as a good benchmark in testing
multifacetedness. A pair of words $a$ and $a^*$ can be chosen based on
the facet or the property of interest with the hope that the
relationship between them is preserved in the vector space.  This will
contribute to a better vector representation of words. 
        
\subsection{Concept Categorization}

An evaluator that is somewhat different from both word similarity and
word analogy is concept categorization. Here, the goal is to split a
given set of words into different categorical subsets of words.  For
example, given the task of separating words into two categories, the
model should be able to categorize words $sandwich$, $tea$, $pasta$,
$water$ into two groups. 

In general, the test can be conducted as follows. First, the
corresponding vector to each word is calculated. Then, a clustering
algorithm (e.g., the $k$ means algorithm) is used to separate the set of
word vectors into $n$ different categories. A performance metric is then
defined based on cluster's purity, where purity refers to whether each
cluster contains concepts from the same or different categories
\cite{baroni2014don}. 
        
By looking at datasets provided for this evaluator, we would like to
point out some challenges. First, the datasets do not have standardized
splits.  Second, no specific clustering methods are defined for this
evaluator. It is important to note that clustering can be computationally
expensive, especially when there are a large amount of words and
categories.  Third, the clustering methods may be unreliable if there
are either uneven distributions of word vectors or no clearly defined
clusters. 
        
Subjectivity is another main issue. As stated by Senel {\em et al.}
\cite{senel2018semantic}, humans can group words by inference using
concepts that word embeddings can gloss over.  Given words $lemon, sun,
banana, blueberry, ocean, iris$. One could group them into yellow
objects ($lemon, sun, banana$) and red objects ($blueberry, ocean,
iris$). Since words can belong to multiple categories, we may argue that
$lemon$, $banana$, $blueberry$, and $iris$ are in the $plant$ category
while $sun$ and $ocean$ are in the $nature$ category. However, due to
the uncompromising nature of the performance metric, there is no
adequate method in evaluating each cluster's quality. 

The property that the sets of words and categories seem to test for is
semantic relation, as words are grouped into concept categories. One
good property of this evaluator is its ability to test for the frequency
effect and the hub-ness problem since it is good at revealing whether
frequent words are clustered together. 
        
\subsection{Outlier Detection}

A relatively new method that evaluates word clustering in vector space
models is outlier detection \cite{outlier}. The goal is to find words
that do not belong to a given group of words. This evaluator tests the
semantic coherence of vector space models, where semantic clusters can
be first identified. There is a clear gold standard for this evaluator
since human performance on this task is extremely high as compared to
word similarity tasks. It is also less subjective. To formalize this
evaluator mathematically, we can take a set of words
\begin{equation}
W = {w_1,w_2,...,w_{n+1}}, 
\end{equation}
where there is one outlier. Next, we take a compactness score of word $w$ as
\begin{equation}
c(w) = \frac{1}{n(n-1)} \sum_{w_i \in W \backslash w} \sum_{w_j \in W
\backslash w, w_j \neq w_i} sim(w_i,w_j). 
\end{equation}
Intuitively, the compactness score of a word is the average of all
pairwise semantic similarities of the words in cluster $W$. The outlier
is the word with the lowest compactness score.  There is less amount of
research on this evaluator as compared with that of word similarity and
word analogy. Yet, it provides a good metric to check whether the
geometry of an embedding space is good. If frequent words are clustered
to form hubs while rarer words are not clustered around the more
frequent words they relate to, the evaluator will not perform well in
this metric. 

There is subjectivity involved in this evaluator as the relationship of
different word groups can be interpreted in different ways. However,
since human perception is often correlated, it may be safe to assume
that this evaluator is objective enough \cite{outlier}. Also, being
similar to the word analogy evaluator, this evaluator relies heavily on
human reasoning and logic. The outliers identified by humans are
strongly influenced by the characteristics of words perceived to be
important. Yet, the recognized patterns might not be immediately clear
to word embedding models. 
    
\subsection{QVEC}

QVEC \cite{qvec} is an intrinsic evaluator that measures the
component-wise correlation between word vectors from a word embedding
model and manually constructed linguistic word vectors in the SemCor
dataset.  These linguistic word vectors are constructed in an attempt to
give well-defined linguistic properties.  QVEC is grounded in the
hypothesis that dimensions in the distributional vectors correspond to
linguistic properties of words. Thus, linear combinations of vector
dimensions produce relevant content. Furthermore, QVEC is a
recall-oriented measure, and highly correlated alignments provide
evaluation and annotations of vector dimensions. Missing information or
noisy dimensions do not signi��cantly affect the score. 

The most prevalent problem with this evaluator is the subjectivity of
man-made linguistic vectors. Current word embedding techniques perform
much better than man-made models as they are based on statistical
relations from data. Having a score based on the correlation between the
word embeddings and the linguistic word vectors may seem to be
counter-intuitive. Thus, the QVEC scores are not very representative of
the performance in downstream tasks. On the other hand, because
linguistic vectors are manually generated, we know exactly which
properties the method is testing for. 

%%%%%%%%%%%%%%%%%%%%%%%%%%%%%%%%%%%%%%%%%%%%%%%%%%%%%%%%%%%%%%%%%%%%%
\begin{table}[htb]
\centering
\caption{Word similarity datasets used in our experiments where pairs 
indicate the number of word pairs in each dataset.}\label{ws_dataset}
\begin{tabular}{|l|l|l|}
\hline
\multicolumn{1}{|c|}{Name} & \multicolumn{1}{c|}{Pairs} & Year \\ \hline
WS-353 \cite{finkelstein2001placing}      & 353                        & 2002      \\ \hline
WS-353-SIM \cite{agirre2009study}                 & 203                & 2009      \\ \hline
WS-353-REL \cite{agirre2009study}                & 252                 & 2009      \\ \hline
MC-30 \cite{miller1991contextual}                      & 30            & 1991      \\ \hline
RG-65 \cite{contextual}                      & 65                      & 1965      \\ \hline
Rare-Word (RW) \cite{luong2013better}                 & 2034           & 2013      \\ \hline
MEN \cite{bruni2014multimodal}                       & 3000            & 2012      \\ \hline
MTurk-287 \cite{radinsky2011word}                  & 287               & 2011      \\ \hline
MTurk-771 \cite{halawi2012large}                 & 771                 & 2012      \\ \hline
YP-130 \cite{turney2001mining}                     & 130               & 2006      \\ \hline
SimLex-999 \cite{hill2015simlex}                 & 999                 & 2014      \\ \hline
Verb-143 \cite{baker2014unsupervised}                   & 143          & 2014      \\ \hline
SimVerb-3500 \cite{gerz2016simverb}                  & 3500            & 2016      \\ \hline
\end{tabular}
\end{table}
%%%%%%%%%%%%%%%%%%%%%%%%%%%%%%%%%%%%%%%%%%%%%%%%%

%%%%%%%%%%%%%%%%%%%%
\begin{table*}[htb]
\caption{Performance comparison ($\times 100$) of six word embedding
baseline models against 13 word similarity datasets.}\label{ws_result}
\centering
\begin{tabular}{cccccccccccccc}
\hline
& \multicolumn{13}{c}{Word Similarity Datasets} \\
\cline{2-14}
   & WS & WS-SIM & WS-REL & MC & RG & RW & MEN & Mturk287 & Mturk771 & YP & SimLex & Verb & SimVerb \\
\hline
SGNS & 71.6 & 78.7 & 62.8 & 81.1 & 79.3 & 46.6 & \textbf{76.1} & 67.3 & \textbf{67.8} & 53.6 & 39.8 & 45.6 & 28.9 \\
CBOW & 64.3 & 74.0 & 53.4 & 74.7 & 81.3 & 43.3 & 72.4 & \textbf{67.4} & 63.6 & 41.6 & 37.2 & 40.9 & 24.5 \\
GloVe & 59.7 & 66.8 & 55.9 & 74.2 & 75.1 & 32.5 & 68.5 & 61.9 & 63.0 & 53.4 & 32.4 & 36.7 & 17.2\\
FastText & 64.8 & 72.1 & 56.4 & 76.3 & 77.3 & 46.6 & 73.0 & 63.0 & 63.0 & 49.0 & 35.2 & 35.0 & 21.9\\
ngram2vec & \textbf{74.2} & \textbf{81.5} & \textbf{67.8} & \textbf{85.7} & 79.5 & 45.0 & 75.1 & 66.5 & 66.5 & 56.4 & \textbf{42.5} & \textbf{47.8} & \textbf{32.1}\\
Dict2vec & 69.4 & 72.8 & 57.3 & 80.5 & \textbf{85.7} & \textbf{49.9} & 73.3 & 60.0 & 65.5 & \textbf{59.6} & 41.7 & 18.9 & 41.7 \\
\hline
\end{tabular}
\end{table*}
%%%%%%%%%%%%%%%%%%%%%%%%%%%%%%%%%%%%%%%%%%%%%%%%%%%%%%%%%%%%%%%%%%%%%

%%%%%%%%%%%%%%%%%%%%
\begin{table*}[htb]
\caption{Performance comparison ($\times 100$) of six word embedding
baseline models against word analogy datasets.}\label{wa_result}
\centering
\begin{tabular}{ccccccccc} \hline
& \multicolumn{8}{c}{Word Analogy Datasets} \\
\cline{2-9}
& \multicolumn{2}{c}{Google} & \multicolumn{2}{c}{Semantic} & \multicolumn{2}{c}{Syntactic} & \multicolumn{2}{c}{MSR}\\

\cline{2-9}
   & Add & Mul & Add & Mul & Add & Mul & Add & Mul \\ \hline
SGNS & \textbf{71.8} & \textbf{73.4} & \textbf{77.6} & \textbf{78.1} & 67.1 & \textbf{69.5} & \textbf{56.7} & \textbf{59.7} \\
CBOW & 70.7 & 70.8 & 74.4 & 74.1 & \textbf{67.6} & 68.1 & 56.2 & 56.8 \\
GloVe & 68.4 & 68.7 & 76.1 & 75.9 & 61.9 & 62.7 & 50.3 & 51.6 \\
FastText & 40.5 & 45.1 & 19.1 & 24.8 & 58.3 & 61.9 & 48.6 & 52.2 \\
ngram2vec & 70.1 & 71.3 & 75.7 & 75.7 & 65.3 & 67.6 & 53.8 & 56.6 \\
Dict2vec & 48.5 & 50.5 & 45.1 & 47.4 & 51.4 & 53.1 & 36.5 & 38.9 \\ \hline
\end{tabular}
\end{table*}
%%%%%%%%%%%%%%%%%%%%%%%%%%%%%%%%%%%%%%%%%%%%%%%%%%%%%%%%%%%%%%%%%%%%%
	
%%%%%%%%%%%%%%%%%%%%%%%%%%%%%%%%%%%%%%%%%%%%%%%%%%%%%%%%%%%%%%%%%%%%%
\begin{table}[htb]
\caption{Performance comparison ($\times 100$) of six word embedding
baseline models against three concept categorization datasets.}\label{cc_result}
\centering
\begin{tabular}{cccc}
\hline
& \multicolumn{3}{c}{Concept Categorization Datasets} \\
\cline{2-4}
& AP & BLESS & BM \\
\cline{2-4}
\hline
SGNS & \textbf{68.2} & 81.0 & \textbf{46.6} \\
CBOW & 65.7 & 74.0 & 45.1  \\
GloVe & 61.4 & \textbf{82.0} & 43.6  \\
FastText & 59.0 & 73.0 & 41.9 \\
ngram2vec & 63.2 & 80.5 & 45.9 \\
Dict2vec & 66.7 & \textbf{82.0} & 46.5 \\ \hline
\end{tabular}
\end{table}
%%%%%%%%%%%%%%%%%%%%%%%%%%%%%%%%%%%%%%%%%%%%%%%%%%%%%%%%%%%%%%%%%%%%%	

%%%%%%%%%%%%%%%%%%%%%%%%%%%%%%%%%%%%%%%%%%%%%%%%%%%%%%%%%%%%%%%%%%%%%
\begin{table}[htb]
\caption{Performance comparison of six word embedding baseline models
against outlier detection datasets.}\label{od_result}
\centering
\begin{tabular}{ccccc} \hline
 & \multicolumn{4}{c}{Outlier Detection Datasets} \\ \cline{2-5}
 & \multicolumn{2}{c}{WordSim-500} & \multicolumn{2}{c}{8-8-8} \\ \cline{2-5}
 & Accuracy & OPP & Accuracy & OPP \\ \hline
SGNS & 11.25 & 83.66 & 57.81 & 84.96 \\
CBOW & 14.02 & 85.33 & 56.25 & 84.38 \\
GloVe & \textbf{15.09} & \textbf{85.74} & 50.0 & 84.77 \\
FastText & 10.68 & 82.16 & 57.81 & 84.38 \\
ngram2vec & 10.64 & 82.83 & 59.38 & \textbf{86.52} \\
Dict2vec & 11.03 & 82.5 & \textbf{60.94} & \textbf{86.52} \\ \hline
\end{tabular}
\end{table}
%%%%%%%%%%%%%%%%%%%%%%%%%%%%%%%%%%%%%%%%%%%%%%%%%%%%%%%%%%%%%%%%%%%%%	

%%%%%%%%%%%%%%%%%%%%%%%%%%%%%%%%%%%%%%%%%%%%%%%%%%%%%%%%%%%%%%%%%%%%%
\begin{table}[htb]
\caption{QVEC performance comparison ($\times 100$) of six word embedding 
baseline models.}\label{QVEC_result}
\centering
\begin{tabular}{cc|cc} \hline
       & QVEC &  & QVEC \\ \hline
SGNS & 50.62 & $FastText$ & 49.20 \\
CBOW & 50.61 & $ngram2vec$ & \textbf{50.83} \\
GloVe & 46.81 & $Dict2vec$ & 48.29 \\ \hline
\end{tabular}
\end{table}
%%%%%%%%%%%%%%%%%%%%%%%%%%%%%%%%%%%%%%%%%%%%%%%%%%%%%%%%%%%%%%%%%%%%		

\section{Experimental Results of Intrinsic Evaluators}\label{sec:exp_IE}

We conduct extensive evaluation experiments on six word embedding models
with intrinsic evaluators in this section. The performance metrics of
consideration include: 1) word similarity, 2) word analogy, 3) concept
categorization, 4) outlier detection and 5) QVEC. 
	
\subsection{Experimental Setup}

We select six word embedding models in the experiments.  They are SGNS,
CBOW, GloVe, FastText, ngram2vec and Dict2vec. For consistency, we
perform training on the same corpus --
wiki2010\footnote{http://nlp.stanford.edu/data/WestburyLab.wikicorp.201004.txt.bz2}.
It is a dataset of medium size (around 6G) without XML tags. After
preprocessing, all special symbols are removed. By choosing a
middle-sized training dataset, we attempt to keep the generality of real
world situations. Some models may perform better when being trained on
larger datasets while others are less dataset dependent.  Here, the same
training dataset is used to fit a more general situation for fair
comparison among different word embedding models. 

For all embedding models, we used their official released toolkit and
default setting for training. For SGNS and CBOW, we used the default
setting provided by the official released
toolkit\footnote{https://code.google.com/archive/p/word2vec/}. GloVe
toolkit is available from their official
website\footnote{https://nlp.stanford.edu/projects/glove/}. For
FastText, we used their
codes\footnote{https://github.com/facebookresearch/fastText}.  Since
FastText uses sub-word as basic units, it can deal with the
out-of-vocabulary (OOV) problem well, which is one of the main
advantages of FastText. Here, to compare the word vector quality only,
we set the vocabulary set for FastText to be the same as other models.
For ngram2vec model\footnote{https://github.com/zhezhaoa/ngram2vec},
because it can be trained over multiple baselines, we chose the best
model reported in their original paper. Finally, codes for Dict2vec can be obtained from
website\footnote{https://github.com/tca19/dict2vec}. The training time
for all models are acceptable (within several hours) using a modern
computer. The threshold for vocabulary is set to 10 for all models. It
means, for words with frequency lower than 10, they are assigned with
the same vectors. 
	
\subsection{Experimental Results}
		
\subsubsection{Word Similarity}

We choose 13 datasets for word similarity evaluation. They are listed in
Table \ref{ws_dataset}. The information of each dataset is provided.
Among the 13 datasets, WS-353, WS-353-SIM, WS-353-REL, Rare-Word are
more popular ones because of their high quality of word pairs.  The
Rare-Word (RW) dataset can be used to test model's ability to learn
words with low frequency. The evaluation result is shown in Table
\ref{ws_result}. We see that SGNS-based models perform better
generally. Note that ngram2vec is an improvement over the SGNS model,
and its performance is the best. Also, The Dict2vec model provides the
best result against the RW dataset. This could be attributed to that
Dict2vec is fine-tuned word vectors based on dictionaries. Since
infrequent words are treated equally with others in dictionaries, the
Dict2vec model is able to give better representation over rare words. 

\subsubsection{Word Analogy}

Two datasets are adopted for the word analogy evaluation task. They are:
1) the Google dataset \cite{word2vec} and 2) the MSR dataset
\cite{3cosadd}. The Google dataset contains 19,544 questions. They are
divided into ``semantic" and ``morpho-syntactic" two categories, each of
which contains 8,869 and 10,675 questions, respectively.  Results for
these two subsets are also reported. The MSR dataset contains 8,000
analogy questions. Both \textit{3CosAdd} and \textit{3CosMul} inference
methods are implemented.  We show the word analogy evaluation results in
Table \ref{wa_result}.  SGNS performs the best. One word set for the
analogy task has four words.  Since ngram2vec considers n-gram models,
the relationship within word sets may not be properly captured.
Dictionaries do not have such word sets and, thus, word analogy is not
well-represented in the word vectors of Dict2vec. Finally, FastText uses
sub-words, its syntactic result is much better than its semantic result. 

\subsubsection{Concept Categorization}

Three datasets are used in concept categorization evaluation.  They are:
1) the AP dataset \cite{almuhareb2006attributes}, 2) the BLESS dataset
\cite{baroni2011we} and 3) the BM dataset \cite{baroni2010strudel}. The
AP dataset contains 402 words that are divided into 21 categories. The
BM dataset is a larger one with 5321 words divided into 56 categories.
Finally, the BLESS dataset consists of 200 words divided into 27
semantic classes.  The results are showed in Table \ref{cc_result}. We
see that the SGNS-based models (including SGNS, ngram2vec and Dict2vec)
perform better than others on all three datasets. 

\subsubsection{Outlier Detection}

We adopt two datasets for the ourlier detection task: 1) the WordSim-500
dataset and 2) the 8-8-8 dataset. The WordSim-500 consists of 500
clusters, where each cluster is represented by a set of 8 words with 5
to 7 outliers \cite{blair2016automated}. The 8-8-8 dataset has 8
clusters, where each cluster is represented by a set of 8 words with 8
outliers \cite{outlier}. Both Accuracy and Outlier Position Percentage
(OPP) are calculated. The results are shown in Table \ref{od_result}.
They are not consistent with each other for the two datasets. For
example, GloVe has the best performance on the WordSim-500 dataset but
its accuracy on the 8-8-8 dataset is the worst. This could be explained
by the properties of these two datasets. We will conduct correlation
study in Sec. \ref{sec:correlation} to shed light on this phenomenon. 

\subsubsection{QVEC}

We use the QVEC toolkit\footnote{https://github.com/ytsvetko/qvec} and
report the sentiment content evaluation result in Table
\ref{QVEC_result}. Among six word models, ngram2vec achieves the best
result while SGNS ranks the second. This is more consistent with other
intrinsic evaluation results described above. 
		
\section{Extrinsic Evaluators}\label{sec:EE}

Based on the definition of extrinsic evaluators, any NLP downstream task
can be chosen as an evaluation method. Here, we present five extrinsic
evaluators: 1) part-of-speech tagging, 2) chunking, 3) named-entity
recognition, 4) sentiment analysis and 5) neural machine translation. 

\subsection{Part-of-speech (POS) Tagging}
	
Part-of-speech (POS) tagging, also called grammar tagging, aims to
assign tags to each input token with its part-of-speech like noun, verb,
adverb, conjunction. Due to the availability of labeled corpora, many
methods can successfully complete this task by either learning
probability distribution through linguistic properties or statistical
machine learning. As low-level linguistic resources, POS tagging can be
used for several purposes such as text indexing and retrieval. 

\subsection{Chunking}
	
The goal of chunking, also called shallow parsing, is to label segments
of a sentence with syntactic constitutes. Each word is first assigned
with one tag indicating its properties such as noun or verb phrases. It
is then used to syntactically grouping words into correlated phrases.
As compared with POS, chunking provides more clues about the structure
of the sentence or phrases in the sentence. 

\subsection{Named-entity Recognition}

The named-entity recognition (NER) task is widely used in natural
language processing. It focuses on reconizing information units such as
names (including person, location and organization) and numeric
expressions (e.g., time and percentage).  Like the POS tagging task, NER
systems use both linguistic grammar-based techniques and statistical
models. A grammar-based system demands lots of efforts on experienced
linguists. In contrast, a statistical-based NER system requires a large
amount of human labeled data for training, and it can achieve higher
precision. Moreover, the current NER systems based on machine learning
are heavily dependent on training data. It may not be robust and cannot
generalize well to different linguistic domains. 

\subsection{Sentiment Analysis}

Sentiment analysis is a particular text classification problem. Usually,
a text fragment is marked with a binary/multi-level label representing
positiveness or negativeness of text's sentiment. An example of this
could be the IMDb dataset by \cite{movie} on whether a given movie
review is positive or negative.  Word phrases are important factor for
final decisions. Negative words such as 'no' or 'not' will totally
reverse the meaning of the whole sentence. Because we are working on
sentence-level or paragraph-level data extraction, word sequence and
parsing plays important role in analyzing sentiment.  Tradition methods
focus more on human-labeled sentence structures. With the development of
machine learning, more statistical and data-driven approaches are
proposed to deal with the sentiment analysis task \cite{ravi2015survey}.
As compared to unlabeled monolingual data, labeled sentiment analysis
data are limited.  Word embedding is commonly used in sentiment analysis
tasks, serving as transferred knowledge extracted from generic large
corpus.  Furthermore, the inference tool is also an important factor,
and it might play a significant role in the final result.  For example,
when conducting sentimental analysis tasks, we may use Bag-of-words,
SVM, LSTM or CNN based on a certain word model. The performance boosts
could be totally different when choosing different inference tools. 

\subsection{Neural Machine Translation (NMT)}

Neural machine translation (NMT) \cite{bahdanau2014neural} refers to a
category of deep-learning-based methods for machine translation. With
large-scale parallel corpus data available, NMT can provide
state-of-the-art results for machine translation and has a large gain
over traditional machine translation methods.  Even with large-scale
parallel data available, domain adaptation is still important to further
improve the performance. Domain adaption methods are able to leverage
monolingual corpus for existing machine translation tasks. As compared
to parallel corpus, monolingual corpus are much larger and they can
provide a model with richer linguistic properties. One representative
domain adaption method is word embedding. This is the reason why NMT can
be used as an extrinsic evaluation task. 
	
\section{Experimental Results of Extrinsic Evaluators}\label{sec:exp_EE}

\subsection{Datasets and Experimental Setup}

\subsubsection{POS Tagging, Chunking and Named Entity Recognition}

By following \cite{collobert2011natural}, three downstream tasks for
sequential labeling are selected in our experiments. The Penn Treebank
(PTB) dataset \cite{marcus1993building}, the chunking of CoNLL'00 share
task dataset \cite{tjong2000introduction} and the NER of CoNLL'03 shared
task dataset \cite{tjong2003introduction} are used for the
part-Of-speech tagging, chunking and named-entity recognition,
respectively. We adopt standard splitting ratios and evaluation criteria
for all three datasets. The details for datasets splitting and
evaluation criteria are shown in Table \ref{POS-C-NER}. 

%%%%%%%%%%%%%%%%%%%%%%%%%%%%%%%%%%%%%%%%%%%%%%%%%%%%%%%%%%%%%%%%%%%
\begin{table}[htb]
\caption{Datasets for POS tagging, Chunking and NER.}\label{POS-C-NER}
\centering
\begin{tabular}{c|c|c|c}
\hline
 Name & Train (\#Tokens) & Test (\#Tokens) & Criteria \\ \hline
PTB & 337,195 & 129,892 & accuracy \\
CoNLL'00 & 211,727 & 47,377 & F-score \\
CoNLL'03 & 203,621 & 46,435 & F-score \\ \hline
\end{tabular}
\end{table}
%%%%%%%%%%%%%%%%%%%%%%%%%%%%%%%%%%%%%%%%%%%%%%%%%%%%%%%%%%%%%%%%%%%%	

For inference tools, we use the simple window-based feed-forward neural
network architecture implemented by \cite{chiu2016intrinsic}. It takes
inputs of five at one time and passes them through a 300-unit hidden
layer, a tanh activation function and a softmax layer before generating
the result. We train each model for 10 epochs using the Adam
optimization with a batch size of 50. 

\subsubsection{Sentiment Analysis}

We choose two sentiment analysis datasets for evaluation: 1) the
Internet Movie Database (IMDb) \cite{movie} and 2) the Stanford
Sentiment Treebank dataset (SST) \cite{socher2013recursive}. IMDb
contains a collection of movie review documents with polarized classes
(positive and negative).  For SST, we split data into three classes:
positive, neutral and negative. Their document formats are different:
IMDb consists several sentences while SST contains only single sentence
per label. The detailed information for each dataset is given in Table
\ref{SA_datasets}. 

%%%%%%%%%%%%%%%%%%%%%%%%%%%%%%%%%%%%%%%%%%%%%%%%%%%%%%%%%%%%%%%%%%%
\begin{table}[htb]
\caption{Sentiment analysis datasets.}\label{SA_datasets}
\centering
\begin{tabular}{c|c|c|c|c} \hline
      & Classes & Train & Validation & Test \\ \hline
SST   & 3 & 8544 & 1101 & 2210 \\
IMDb  & 2 & 17500 & 7500 & 25000 \\ \hline
\end{tabular}
\end{table}
%%%%%%%%%%%%%%%%%%%%%%%%%%%%%%%%%%%%%%%%%%%%%%%%%%%%%%%%%%%%%%%%%%%%	

To cover most sentimental analysis inference tools, we test the task
using Bi-LSTM and CNN. We choose 2-layer Bi-LSTM with 256 hidden
dimensions. The adopted CNN has 3 layers with 100 filters per layer of
size [3, 4, 5], respectively. Particularly, the embedding layer for all
models are fixed during training. All models are trained for 5 epochs
using the Adam optimization with 0.0001 learning rate.  

\subsubsection{Neural Machine Translation}

As compared with sentiment analysis, neural machine translation (NMT) is
a more challenging task since it demands a larger network and more
training data. We use the same encoder-decoder architecture as that in
\cite{klein2017opennmt}. The Europarl v8 \cite{koehn2005europarl}
dataset is used as training corpora. The task is English-French
translation. For French word embedding, a pre-trained FastText word
embedding model\footnote{https://github.com/facebookresearch/fastText/blob/master/pretrained-vectors.md}
is utilized. As to the hyper-parameter setting, we use a single layer
bidirectional-LSTM of 500 dimensions for both the encoder and the
decoder.  Both embedding layers for the encoder and the decoder are
fixed during the training process. The batch size is 30 and the total
training iteration is 100,000. 

\subsection{Experimental Results and Discussion}

Experimental results of the above-mentioned five extrinsic evaluators
are shown in Table \ref{EE_results}. Generally speaking, both SGNS and
ngram2vec perform well in POS tagging, chunking and NER tasks.
Actually, the performance differences of all evaluators are small in
these three tasks.  As to the sentimental analysis, their is no obvious
winner with the CNN inference tool. The performance gaps become larger
using the Bi-LSTM inference tool, and we see that Dict2vec and FastText
perform the worst. Based on these results, we observe that there exist
two different factors affecting the sentiment analysis results: datasets
and inference tools. For different datasets with the same inference
tool, the performance can be different because of different linguistic
properties of datasets. On the other hand, different inference tools may
favor different embedding models against the same dataset since
inference tools extract the information from word models in their own
manner. For example, Bi-LSTM focuses on long range dependency while CNN
treats each token more or less equally. 

%%%%%%%%%%%%%%%%%%%%%%%%%%%%%%%%%%%%%%%%%%%%%%%%%%%%%%%%%%%%%%%%%%%%
\begin{table*}[htb]
\caption{Extrinsic evaluation results.}\label{EE_results}
\centering
\begin{tabular}{c|c|c|c|c|c|c|c|c} \hline
& \multirow{2}{*}{POS}  & \multirow{2}{*}{Chunking} & \multirow{2}{*}{NER} & \multicolumn{2}{c|}{SA(IMDb)} & 
\multicolumn{2}{c|}{SA(SST)} & NMT\\ \cline{5-9}
 &  & & & Bi-LSTM  & CNN & Bi-LSTM & CNN & Perplexity\\ \hline
SGNS  & \textbf{94.54} & 88.21 & 87.12 & 85.36 & 88.78 & 64.08  & \textbf{66.93} & 79.14 \\
CBOW  & 93.79 & 84.91 & 83.83 & \textbf{86.93}  & 85.88 & 65.63  & 65.06 & 102.33 \\
GloVe & 93.32 & 84.11 & 85.3 & 70.41  & 87.56 & 65.16  & 65.15 & 84.20 \\
FastText  & 94.36 & 87.96 & 87.10 & 73.97  & 83.69 & 50.01 & 63.25 & 82.60 \\
ngram2vec & 94.11 & \textbf{88.74} & \textbf{87.33} & 79.32 & \textbf{89.29} & \textbf{66.27} & 66.45 & \textbf{77.79} \\
Dict2vec & 93.61 & 86.54 & 86.82 & 62.71 & 88.94  & 62.75 & 66.09 & 78.84 \\ \hline
\end{tabular}
\end{table*}
%%%%%%%%%%%%%%%%%%%%%%%%%%%%%%%%%%%%%%%%%%%%%%%%%%%%%%%%%%%%%%%%%%%%

Perplexity is used to evaluate the NMT task.  It indicates variability
of a prediction model. Lower perplexity corresponds to lower entropy
and, thus, better performance.  We separate 20,000 sentences from the
same corpora to generate testing data and report testing perplexity for
the NMT task in Table \ref{EE_results}.  As shown in the table,
ngram2vec, Dict2vec and SGNS are the top three word models for the NMT
task, which is consistent with the word similarity evaluation results. 

We conclude from Table \ref{EE_results} that SGNS-based models including
SGNS, ngram2vec and dict2vec tend to work better than other models.
However, one drawback of ngram2vec is that it takes more time in
processing n-gram data for training. GloVe and FastText are popular in
the research community since their pre-trained models are easy to
download. We also compared results using pre-trained GloVe and FastText
models. Although they are both trained on larger datasets and properly
find-tuned, they do not provide better results in our evaluation tasks. 

\section{Consistency Study via Correlation Analysis}\label{sec:correlation}

We conduct consistency study of extrinsic and intrinsic evaluators using
the Pearson correlation ($\rho$) analysis \cite{benesty2009pearson}.
Besides the six word models described above, we add two more pre-trained
models of GloVe and FastText to make the total model number eight.
Furthermore, we apply the variance normalization technique
\cite{wang2018post} to the eight models to yield eight more models.
Consequently, we have a collection of sixteen word models. 

\begin{figure*}[htb]
\centering
\includegraphics[width=\linewidth]{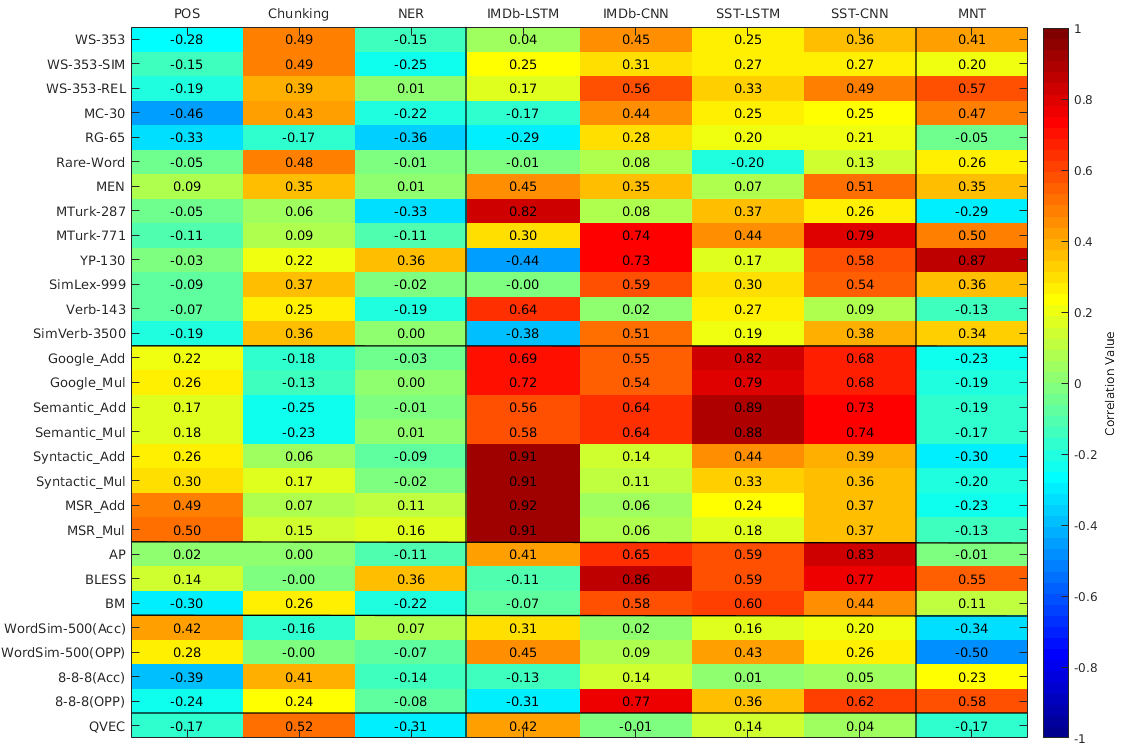}
\caption{Pearson's correlation between intrinsic and extrinsic
evaluator, where the x-axis shows extrinsic evaluators while the y-axis
indicates intrinsic evaluators. The warm indicates the positive
correlation while the cool color indicates the negative correlation.}
\label{fig_correlation}
\end{figure*}

Fig. \ref{fig_correlation} shows the Pearson correlation of each
intrinsic and extrinsic evaluation pair of these sixteen models.  For
example, the entry of the first row and the first column is the Pearson
correlation value of WS-353 (an intrinsic evaluator) and POS (an
extrinsic evaluator) of sixteen word models (i.e. 16 evaluation data
pairs). Note also that we add a negative sign to the correlation value
of NMT perplexity since lower perplexity is better. 

%%%%%%%%%%%%%%%%%%%%%%%%%%%%%%%%%%%%%%%%%%%%%%%%%%%%%%%%%%%%%%%%%%%%
%%%%%%%%%%%%%%%%%%%%%%%%%%%%%%%%%%%%%%%%%%%%%%%%%%%%%%%%%%%%%%%%%%%%

\subsection{Consistency of Intrinsic Evaluators}

\begin{itemize}
\item \textit{Word Similarity}

All embedding models are tested over 13 evaluation datasets and the
results are shown in the top 13 rows. We see from the correlation result
that larger datasets tend to give more reliable and consistent
evaluation result. Among all datasets, WS-353, WS-353-SIM, WS-353-REL,
MTrurk-771, SimLex-999 and SimVerb-3500 are recommended to serve as
generic evaluation datasets. Although datasets like MC-30 and RG-65 also
provide us with reasonable results, their correlation results are not as
consistent as others. This may be attributed to the limited amount of
testing samples with only dozens of testing word pairs.  The Rare-Word
(RW) dataset is a special one that focuses on low-frequency words and
gains popularity recently. Yet, based on the correlation study, the RW
dataset is not as effective as expected. Infrequent words may not play
an important role in all extrinsic evaluation tasks. This is why
infrequent words are often set to the same vector.  The Rare-Word
dataset can be excluded for general purpose evaluation unless there is a
specific application demanding rare words modeling. 

\item \textit{Word Analogy}

The word analogy results are shown from the 14th row to the 21st row in
the figure.  Among four word analogy datasets (i.e. Google, Google
Semantic, Google Syntactic and MSR), Google and Google Semantic are more
effective. It does not make much difference in the final correlation
study using either the 3CosAdd or the 3CosMul compuation. Google
Syntactic is not effective since the morphology of words does not
contain as much information as semantic meanings.  Thus, although the
FastText model performs well in morphology testing based on the average
of sub-words, it correlation analysis is worse than other models.  In
general, word analogy provides most reliable correlation results and
has the highest correlation with the sentiment analysis task. 

\item \textit{Concept Categorization}

All three datasets (i.e., AP, BLESS and BM) for concept categorization
perform well. By categorizing words into different groups, concept
categorization focuses on semantic clusters. It appears that models that
are good at dividing words into semantic collections are more effective
in downstream NLP tasks. 

\item \textit{Outlier Detection}

Two datasets (i.e., WordSim-500 and 8-8-8) are used for outlier
detection.  In general, outlier detection is not a good evaluation
method.  Although it tests semantic clusters to some extent, outlier
detection is less direct as compared to concept categorization. Also,
from the dataset point of view, the size of the 8-8-8 dataset is too
small while the WordSim-500 dataset contains too many infrequent words
in the clusters. This explains why the accuracy for WordSim-500 is low
(around 10-20\%). When there are larger and more reliable datasets
available, we expect the outlier detection task to have better
performance in word embedding evaluation. 

\item \textit{QVEC}

QVEC is not a good evaluator due to its inherit properties. It attempts
to compute the correlation with lexicon-resource based word vectors.
Yet, the quality of lexicon-resource based word vectors is to poor to
provide a reliable rule. If we can find a more reliable rule, the QVEC
evaluator will perform better. 

\end{itemize}

Based on the above discussion, we conclude that word similarity, word
analogy and concept categorization are more effective intrinsic
evaluators. Different datasets lead to different performance. In
general, larger datasets tend to give better and more reliable results.
Intrinsic evaluators may perform very differently for different
downstream tasks.  Thus, when we test a new word embedding model, all
three intrinsic evaluators should be used and considered jointly. 

\subsection{Consistency of Extrinsic Evaluators}

For POS tagging, chunking and NER, none of intrinsic evaluators provide
high correlation. Their performance depend on their capability in
sequential information extraction.  Thus, word meaning plays a
subsidiary role in all these tasks.  Sentiment analysis is a
dimensionality reduction procedure. It focuses more on combination of
word meaning. Thus, it has stronger correlation with the properties that
the word analogy evaluator is testing.  Finally, NMT is
sentence-to-sentence conversion, and the mapping between word pairs is
more helpful in translation tasks. Thus, the word similarity evaluator
has a stronger correlation with the NMT task. We should also point out
that some unsupervised machine translation tasks focus on word pairs
\cite{artetxe2017learning,artetxe2017unsupervised}. This shows the
significance of word pair correspondence in NMT. 

\section{Conclusion and Future Work}\label{sec:future}

In this work, we provided in-depth discussion of intrinsic and extrinsic
evaluations on many word embedding models, showed extensive experimental
results and explained the observed phenomema. Our study offers a
valuable guidance in selecting suitable evaluation methods for different
application tasks.  There are many factors affecting word embedding
quality. Furthermore, there are still no perfect evaluation methods
testing the word subspace for linguistic relationships because it is
difficult to understand exactly how the embedding spaces encode
linguistic relations. For this reason, we expect more work to be done in
developing better metrics for evaluation on the overall quality of a
word model.  Such metrics must be computationally efficient while
having a high correlation with extrinsic evaluation test scores. The
crux of this problem lies in decoding how the word subspace encodes
linguistic relations and the quality of these relations. 

We would like to point out that linguistic relations and properties
captured by word embedding models are different from how humans learn
languages. For humans, a language encompasses many different avenues
e.g., a sense of reasoning, cultural differences, contextual
implications and many others. Thus, a language is filled with subjective
complications that interfere with objective goals of models.  In
contrast, word embedding models perform well in specific applied tasks.
They have triumphed over the work of linguists in creating taxonomic
structures and other manually generated representations.  Yet, different
datasets and different models are used for different specific tasks.  

We do not see a word embedding model that consistently performs well in
all tasks.  The design of a more universal word embedding model is
challenging. To generate word models that are good at solving specific
tasks, task-specific data can be fed into a model for training. Feeding
a large amount of generic data can be inefficient and even hurt the
performance of a word model since different task-specific data can lead
to contending results. It is still not clear what is the proper balance
between the two design methodologies. 

\bibliographystyle{IEEEtran}
\bibliography{IEEEabrv,IEEEexample}

\begin{thebibliography}{10}
\providecommand{\url}[1]{#1}
\csname url@samestyle\endcsname
\providecommand{\newblock}{\relax}
\providecommand{\bibinfo}[2]{#2}
\providecommand{\BIBentrySTDinterwordspacing}{\spaceskip=0pt\relax}
\providecommand{\BIBentryALTinterwordstretchfactor}{4}
\providecommand{\BIBentryALTinterwordspacing}{\spaceskip=\fontdimen2\font plus
\BIBentryALTinterwordstretchfactor\fontdimen3\font minus
  \fontdimen4\font\relax}
\providecommand{\BIBforeignlanguage}[2]{{%
\expandafter\ifx\csname l@#1\endcsname\relax
\typeout{** WARNING: IEEEtran.bst: No hyphenation pattern has been}%
\typeout{** loaded for the language `#1'. Using the pattern for}%
\typeout{** the default language instead.}%
\else
\language=\csname l@#1\endcsname
\fi
#2}}
\providecommand{\BIBdecl}{\relax}
\BIBdecl

\bibitem{yu2018refining}
L.-C. Yu, J.~Wang, K.~R. Lai, and X.~Zhang, ``Refining word embeddings using
  intensity scores for sentiment analysis,'' \emph{IEEE/ACM Transactions on
  Audio, Speech and Language Processing (TASLP)}, vol.~26, no.~3, pp. 671--681,
  2018.

\bibitem{schutze2008introduction}
H.~Sch{\"u}tze, C.~D. Manning, and P.~Raghavan, \emph{Introduction to
  information retrieval}.\hskip 1em plus 0.5em minus 0.4em\relax Cambridge
  University Press, 2008, vol.~39.

\bibitem{chen2015distributed}
W.~Chen, M.~Zhang, and Y.~Zhang, ``Distributed feature representations for
  dependency parsing,'' \emph{IEEE Transactions on Audio, Speech, and Language
  Processing}, vol.~23, no.~3, pp. 451--460, 2015.

\bibitem{ouchi2016transition}
H.~Ouchi, K.~Duh, H.~Shindo, and Y.~Matsumoto, ``Transition-based dependency
  parsing exploiting supertags,'' \emph{IEEE/ACM Transactions on Audio, Speech,
  and Language Processing}, vol.~24, no.~11, pp. 2059--2068, 2016.

\bibitem{shen2014dependency}
M.~Shen, D.~Kawahara, and S.~Kurohashi, ``Dependency parse reranking with rich
  subtree features,'' \emph{IEEE/ACM Transactions on Audio, Speech, and
  Language Processing}, vol.~22, no.~7, pp. 1208--1218, 2014.

\bibitem{zhou2016learning}
G.~Zhou, Z.~Xie, T.~He, J.~Zhao, and X.~T. Hu, ``Learning the multilingual
  translation representations for question retrieval in community question
  answering via non-negative matrix factorization,'' \emph{IEEE/ACM
  Transactions on Audio, Speech and Language Processing (TASLP)}, vol.~24,
  no.~7, pp. 1305--1314, 2016.

\bibitem{hao2017end}
Y.~Hao, Y.~Zhang, K.~Liu, S.~He, Z.~Liu, H.~Wu, and J.~Zhao, ``An end-to-end
  model for question answering over knowledge base with cross-attention
  combining global knowledge,'' in \emph{Proceedings of the 55th Annual Meeting
  of the Association for Computational Linguistics (Volume 1: Long Papers)},
  vol.~1, 2017, pp. 221--231.

\bibitem{zhang2017context}
B.~Zhang, D.~Xiong, J.~Su, and H.~Duan, ``A context-aware recurrent encoder for
  neural machine translation,'' \emph{IEEE/ACM Transactions on Audio, Speech
  and Language Processing (TASLP)}, vol.~25, no.~12, pp. 2424--2432, 2017.

\bibitem{chen2018neural}
K.~Chen, T.~Zhao, M.~Yang, L.~Liu, A.~Tamura, R.~Wang, M.~Utiyama, and
  E.~Sumita, ``A neural approach to source dependence based context model for
  statistical machine translation,'' \emph{IEEE/ACM Transactions on Audio,
  Speech and Language Processing (TASLP)}, vol.~26, no.~2, pp. 266--280, 2018.

\bibitem{survey}
\BIBentryALTinterwordspacing
A.~Bakarov, ``A survey of word embeddings evaluation methods,'' \emph{CoRR},
  vol. abs/1801.09536, 2018. [Online]. Available:
  \url{http://arxiv.org/abs/1801.09536}
\BIBentrySTDinterwordspacing

\bibitem{li2014joint}
Z.~Li, M.~Zhang, W.~Che, T.~Liu, and W.~Chen, ``Joint optimization for chinese
  pos tagging and dependency parsing,'' \emph{IEEE/ACM Transactions on Audio,
  Speech and Language Processing (TASLP)}, vol.~22, no.~1, pp. 274--286, 2014.

\bibitem{xu2018cross}
J.~Xu, X.~Sun, H.~He, X.~Ren, and S.~Li, ``Cross-domain and semi-supervised
  named entity recognition in chinese social media: A unified model,''
  \emph{IEEE/ACM Transactions on Audio, Speech, and Language Processing}, 2018.

\bibitem{ravi2015survey}
K.~Ravi and V.~Ravi, ``A survey on opinion mining and sentiment analysis:
  tasks, approaches and applications,'' \emph{Knowledge-Based Systems},
  vol.~89, pp. 14--46, 2015.

\bibitem{bahdanau2014neural}
D.~Bahdanau, K.~Cho, and Y.~Bengio, ``Neural machine translation by jointly
  learning to align and translate,'' \emph{arXiv preprint arXiv:1409.0473},
  2014.

\bibitem{eval_methods}
T.~Schnabel, I.~Labutov, D.~Mimno, and T.~Joachims, ``Evaluation methods for
  unsupervised word embeddings,'' in \emph{Proceedings of the 2015 Conference
  on Empirical Methods in Natural Language Processing}, 2015, pp. 298--307.

\bibitem{chiu2016intrinsic}
B.~Chiu, A.~Korhonen, and S.~Pyysalo, ``Intrinsic evaluation of word vectors
  fails to predict extrinsic performance,'' in \emph{Proceedings of the 1st
  Workshop on Evaluating Vector-Space Representations for NLP}, 2016, pp. 1--6.

\bibitem{qiu2018revisiting}
Y.~Qiu, H.~Li, S.~Li, Y.~Jiang, R.~Hu, and L.~Yang, ``Revisiting correlations
  between intrinsic and extrinsic evaluations of word embeddings,'' in
  \emph{Chinese Computational Linguistics and Natural Language Processing Based
  on Naturally Annotated Big Data}.\hskip 1em plus 0.5em minus 0.4em\relax
  Springer, 2018, pp. 209--221.

\bibitem{nnlm}
\BIBentryALTinterwordspacing
Y.~Bengio, R.~Ducharme, P.~Vincent, and C.~Janvin, ``A neural probabilistic
  language model,'' \emph{Journal of Machine Learning Research}, vol.~3, pp.
  1137--1155, 2003. [Online]. Available:
  \url{http://www.jmlr.org/papers/v3/bengio03a.html}
\BIBentrySTDinterwordspacing

\bibitem{word2vec}
\BIBentryALTinterwordspacing
T.~Mikolov, K.~Chen, G.~Corrado, and J.~Dean, ``Efficient estimation of word
  representations in vector space,'' \emph{CoRR}, vol. abs/1301.3781, 2013.
  [Online]. Available: \url{http://arxiv.org/abs/1301.3781}
\BIBentrySTDinterwordspacing

\bibitem{softmax}
T.~Mikolov, I.~Sutskever, K.~Chen, G.~S. Corrado, and J.~Dean, ``Distributed
  representations of words and phrases and their compositionality,'' in
  \emph{Advances in neural information processing systems}, 2013, pp.
  3111--3119.

\bibitem{glove}
J.~Pennington, R.~Socher, and C.~Manning, ``Glove: Global vectors for word
  representation,'' in \emph{Proceedings of the 2014 conference on empirical
  methods in natural language processing (EMNLP)}, 2014, pp. 1532--1543.

\bibitem{fasttext}
\BIBentryALTinterwordspacing
P.~Bojanowski, E.~Grave, A.~Joulin, and T.~Mikolov, ``Enriching word vectors
  with subword information,'' \emph{CoRR}, vol. abs/1607.04606, 2016. [Online].
  Available: \url{http://arxiv.org/abs/1607.04606}
\BIBentrySTDinterwordspacing

\bibitem{zhao2017ngram2vec}
Z.~Zhao, T.~Liu, S.~Li, B.~Li, and X.~Du, ``Ngram2vec: Learning improved word
  representations from ngram co-occurrence statistics,'' in \emph{Proceedings
  of the 2017 Conference on Empirical Methods in Natural Language Processing},
  2017, pp. 244--253.

\bibitem{tissier2017dict2vec}
J.~Tissier, C.~Gravier, and A.~Habrard, ``Dict2vec: Learning word embeddings
  using lexical dictionaries,'' in \emph{Conference on Empirical Methods in
  Natural Language Processing (EMNLP 2017)}, 2017, pp. 254--263.

\bibitem{deep}
M.~E. Peters, M.~Neumann, M.~Iyyer, M.~Gardner, C.~Clark, K.~Lee, and
  L.~Zettlemoyer, ``Deep contextualized word representations,'' \emph{arXiv
  preprint arXiv:1802.05365}, 2018.

\bibitem{devlin2018bert}
J.~Devlin, M.-W. Chang, K.~Lee, and K.~Toutanova, ``Bert: Pre-training of deep
  bidirectional transformers for language understanding,'' \emph{arXiv preprint
  arXiv:1810.04805}, 2018.

\bibitem{radford2018improving}
A.~Radford, K.~Narasimhan, T.~Salimans, and I.~Sutskever, ``Improving language
  understanding by generative pre-training,'' \emph{Technical report, OpenAI},
  2018.

\bibitem{props}
Y.~Yaghoobzadeh and H.~Sch{\"u}tze, ``Intrinsic subspace evaluation of word
  embedding representations,'' \emph{arXiv preprint arXiv:1606.07902}, 2016.

\bibitem{reliability}
J.~Hellrich and U.~Hahn, ``Don’t get fooled by word embeddings: better watch
  their neighborhood,'' in \emph{Digital Humanities 2017—Conference Abstracts
  of the 2017 Conference of the Alliance of Digital Humanities Organizations
  (ADHO). Montr{\'e}al, Quebec, Canada}, 2017, pp. 250--252.

\bibitem{goodgeo}
A.~Gladkova and A.~Drozd, ``Intrinsic evaluations of word embeddings: What can
  we do better?'' in \emph{Proceedings of the 1st Workshop on Evaluating
  Vector-Space Representations for NLP}, 2016, pp. 36--42.

\bibitem{statsig}
W.~Shalaby and W.~Zadrozny, ``Measuring semantic relatedness using mined
  semantic analysis,'' \emph{CoRR, abs/1512.03465}, 2015.

\bibitem{ws}
M.~Faruqui, Y.~Tsvetkov, P.~Rastogi, and C.~Dyer, ``Problems with evaluation of
  word embeddings using word similarity tasks,'' \emph{arXiv preprint
  arXiv:1605.02276}, 2016.

\bibitem{3cosadd}
T.~Mikolov, W.-t. Yih, and G.~Zweig, ``Linguistic regularities in continuous
  space word representations,'' in \emph{Proceedings of the 2013 Conference of
  the North American Chapter of the Association for Computational Linguistics:
  Human Language Technologies}, 2013, pp. 746--751.

\bibitem{pairdir}
O.~Levy and Y.~Goldberg, ``Linguistic regularities in sparse and explicit word
  representations,'' in \emph{Proceedings of the eighteenth conference on
  computational natural language learning}, 2014, pp. 171--180.

\bibitem{analogy_issues}
A.~Rogers, A.~Drozd, and B.~Li, ``The (too many) problems of analogical
  reasoning with word vectors,'' in \emph{Proceedings of the 6th Joint
  Conference on Lexical and Computational Semantics (* SEM 2017)}, 2017, pp.
  135--148.

\bibitem{baroni2014don}
M.~Baroni, G.~Dinu, and G.~Kruszewski, ``Don't count, predict! a systematic
  comparison of context-counting vs. context-predicting semantic vectors,'' in
  \emph{Proceedings of the 52nd Annual Meeting of the Association for
  Computational Linguistics (Volume 1: Long Papers)}, vol.~1, 2014, pp.
  238--247.

\bibitem{senel2018semantic}
L.~K. Senel, I.~Utlu, V.~Yucesoy, A.~Koc, and T.~Cukur, ``Semantic structure
  and interpretability of word embeddings,'' \emph{IEEE/ACM Transactions on
  Audio, Speech, and Language Processing}, 2018.

\bibitem{outlier}
J.~Camacho-Collados and R.~Navigli, ``Find the word that does not belong: A
  framework for an intrinsic evaluation of word vector representations,'' in
  \emph{Proceedings of the 1st Workshop on Evaluating Vector-Space
  Representations for NLP}, 2016, pp. 43--50.

\bibitem{qvec}
Y.~Tsvetkov, M.~Faruqui, W.~Ling, G.~Lample, and C.~Dyer, ``Evaluation of word
  vector representations by subspace alignment,'' in \emph{Proceedings of the
  2015 Conference on Empirical Methods in Natural Language Processing}, 2015,
  pp. 2049--2054.

\bibitem{finkelstein2001placing}
L.~Finkelstein, E.~Gabrilovich, Y.~Matias, E.~Rivlin, Z.~Solan, G.~Wolfman, and
  E.~Ruppin, ``Placing search in context: The concept revisited,'' in
  \emph{Proceedings of the 10th international conference on World Wide
  Web}.\hskip 1em plus 0.5em minus 0.4em\relax ACM, 2001, pp. 406--414.

\bibitem{agirre2009study}
E.~Agirre, E.~Alfonseca, K.~Hall, J.~Kravalova, M.~Pa{\c{s}}ca, and A.~Soroa,
  ``A study on similarity and relatedness using distributional and
  wordnet-based approaches,'' in \emph{Proceedings of Human Language
  Technologies: The 2009 Annual Conference of the North American Chapter of the
  Association for Computational Linguistics}.\hskip 1em plus 0.5em minus
  0.4em\relax Association for Computational Linguistics, 2009, pp. 19--27.

\bibitem{miller1991contextual}
G.~A. Miller and W.~G. Charles, ``Contextual correlates of semantic
  similarity,'' \emph{Language and cognitive processes}, vol.~6, no.~1, pp.
  1--28, 1991.

\bibitem{contextual}
H.~Rubenstein and J.~B. Goodenough, ``Contextual correlates of synonymy,''
  \emph{Communications of the ACM}, vol.~8, no.~10, pp. 627--633, 1965.

\bibitem{luong2013better}
T.~Luong, R.~Socher, and C.~Manning, ``Better word representations with
  recursive neural networks for morphology,'' in \emph{Proceedings of the
  Seventeenth Conference on Computational Natural Language Learning}, 2013, pp.
  104--113.

\bibitem{bruni2014multimodal}
E.~Bruni, N.-K. Tran, and M.~Baroni, ``Multimodal distributional semantics,''
  \emph{Journal of Artificial Intelligence Research}, vol.~49, pp. 1--47, 2014.

\bibitem{radinsky2011word}
K.~Radinsky, E.~Agichtein, E.~Gabrilovich, and S.~Markovitch, ``A word at a
  time: computing word relatedness using temporal semantic analysis,'' in
  \emph{Proceedings of the 20th international conference on World wide
  web}.\hskip 1em plus 0.5em minus 0.4em\relax ACM, 2011, pp. 337--346.

\bibitem{halawi2012large}
G.~Halawi, G.~Dror, E.~Gabrilovich, and Y.~Koren, ``Large-scale learning of
  word relatedness with constraints,'' in \emph{Proceedings of the 18th ACM
  SIGKDD international conference on Knowledge discovery and data
  mining}.\hskip 1em plus 0.5em minus 0.4em\relax ACM, 2012, pp. 1406--1414.

\bibitem{turney2001mining}
P.~D. Turney, ``Mining the web for synonyms: Pmi-ir versus lsa on toefl,'' in
  \emph{European Conference on Machine Learning}.\hskip 1em plus 0.5em minus
  0.4em\relax Springer, 2001, pp. 491--502.

\bibitem{hill2015simlex}
F.~Hill, R.~Reichart, and A.~Korhonen, ``Simlex-999: Evaluating semantic models
  with (genuine) similarity estimation,'' \emph{Computational Linguistics},
  vol.~41, no.~4, pp. 665--695, 2015.

\bibitem{baker2014unsupervised}
S.~Baker, R.~Reichart, and A.~Korhonen, ``An unsupervised model for instance
  level subcategorization acquisition,'' in \emph{Proceedings of the 2014
  Conference on Empirical Methods in Natural Language Processing (EMNLP)},
  2014, pp. 278--289.

\bibitem{gerz2016simverb}
D.~Gerz, I.~Vuli{\'c}, F.~Hill, R.~Reichart, and A.~Korhonen, ``Simverb-3500: A
  large-scale evaluation set of verb similarity,'' \emph{arXiv preprint
  arXiv:1608.00869}, 2016.

\bibitem{almuhareb2006attributes}
A.~Almuhareb, ``Attributes in lexical acquisition,'' Ph.D. dissertation,
  University of Essex, 2006.

\bibitem{baroni2011we}
M.~Baroni and A.~Lenci, ``How we blessed distributional semantic evaluation,''
  in \emph{Proceedings of the GEMS 2011 Workshop on GEometrical Models of
  Natural Language Semantics}.\hskip 1em plus 0.5em minus 0.4em\relax
  Association for Computational Linguistics, 2011, pp. 1--10.

\bibitem{baroni2010strudel}
M.~Baroni, B.~Murphy, E.~Barbu, and M.~Poesio, ``Strudel: A corpus-based
  semantic model based on properties and types,'' \emph{Cognitive science},
  vol.~34, no.~2, pp. 222--254, 2010.

\bibitem{blair2016automated}
P.~Blair, Y.~Merhav, and J.~Barry, ``Automated generation of multilingual
  clusters for the evaluation of distributed representations,'' \emph{arXiv
  preprint arXiv:1611.01547}, 2016.

\bibitem{movie}
A.~L. Maas, R.~E. Daly, P.~T. Pham, D.~Huang, A.~Y. Ng, and C.~Potts,
  ``Learning word vectors for sentiment analysis,'' in \emph{Proceedings of the
  49th annual meeting of the association for computational linguistics: Human
  language technologies-volume 1}.\hskip 1em plus 0.5em minus 0.4em\relax
  Association for Computational Linguistics, 2011, pp. 142--150.

\bibitem{collobert2011natural}
R.~Collobert, J.~Weston, L.~Bottou, M.~Karlen, K.~Kavukcuoglu, and P.~Kuksa,
  ``Natural language processing (almost) from scratch,'' \emph{Journal of
  Machine Learning Research}, vol.~12, no. Aug, pp. 2493--2537, 2011.

\bibitem{marcus1993building}
M.~P. Marcus, M.~A. Marcinkiewicz, and B.~Santorini, ``Building a large
  annotated corpus of english: The penn treebank,'' \emph{Computational
  linguistics}, vol.~19, no.~2, pp. 313--330, 1993.

\bibitem{tjong2000introduction}
E.~F. Tjong Kim~Sang and S.~Buchholz, ``Introduction to the conll-2000 shared
  task: Chunking,'' in \emph{Proceedings of the 2nd workshop on Learning
  language in logic and the 4th conference on Computational natural language
  learning-Volume 7}.\hskip 1em plus 0.5em minus 0.4em\relax Association for
  Computational Linguistics, 2000, pp. 127--132.

\bibitem{tjong2003introduction}
E.~F. Tjong Kim~Sang and F.~De~Meulder, ``Introduction to the conll-2003 shared
  task: Language-independent named entity recognition,'' in \emph{Proceedings
  of the seventh conference on Natural language learning at HLT-NAACL
  2003-Volume 4}.\hskip 1em plus 0.5em minus 0.4em\relax Association for
  Computational Linguistics, 2003, pp. 142--147.

\bibitem{socher2013recursive}
R.~Socher, A.~Perelygin, J.~Wu, J.~Chuang, C.~D. Manning, A.~Ng, and C.~Potts,
  ``Recursive deep models for semantic compositionality over a sentiment
  treebank,'' in \emph{Proceedings of the 2013 conference on empirical methods
  in natural language processing}, 2013, pp. 1631--1642.

\bibitem{klein2017opennmt}
G.~Klein, Y.~Kim, Y.~Deng, J.~Senellart, and A.~M. Rush, ``Opennmt: Open-source
  toolkit for neural machine translation,'' \emph{arXiv preprint
  arXiv:1701.02810}, 2017.

\bibitem{koehn2005europarl}
P.~Koehn, ``Europarl: A parallel corpus for statistical machine translation,''
  in \emph{MT summit}, vol.~5, 2005, pp. 79--86.

\bibitem{benesty2009pearson}
J.~Benesty, J.~Chen, Y.~Huang, and I.~Cohen, ``Pearson correlation
  coefficient,'' in \emph{Noise reduction in speech processing}.\hskip 1em plus
  0.5em minus 0.4em\relax Springer, 2009, pp. 1--4.

\bibitem{wang2018post}
B.~Wang, F.~Chen, A.~Wang, and C.-C.~J. Kuo, ``Post-processing of word
  representations via variance normalization and dynamic embedding,''
  \emph{arXiv preprint arXiv:1808.06305}, 2018.

\bibitem{artetxe2017learning}
M.~Artetxe, G.~Labaka, and E.~Agirre, ``Learning bilingual word embeddings with
  (almost) no bilingual data,'' in \emph{Proceedings of the 55th Annual Meeting
  of the Association for Computational Linguistics (Volume 1: Long Papers)},
  vol.~1, 2017, pp. 451--462.

\bibitem{artetxe2017unsupervised}
M.~Artetxe, G.~Labaka, E.~Agirre, and K.~Cho, ``Unsupervised neural machine
  translation,'' \emph{arXiv preprint arXiv:1710.11041}, 2017.

\end{thebibliography}

\end{document}